\documentclass[letterpaper, 10 pt, conference]{ieeeconf}  
\IEEEoverridecommandlockouts                           
\usepackage{graphics} 
\usepackage{epsfig} 
\usepackage{mathptmx} 
\usepackage{times} 
\usepackage{amsmath} 
\usepackage{amssymb}  
\usepackage{xcolor}
\usepackage{array, multirow,graphicx}
\usepackage{float}
\usepackage{graphicx}
\usepackage{subfigure}
\let\labelindent\relax
\usepackage{enumitem}
\usepackage{lipsum}
\usepackage{url}
\usepackage{hyperref}
\hypersetup{
    colorlinks=true,
    citecolor=black,
    filecolor=black,
    urlcolor=gray,
    breaklinks=true,
}

\makeatletter
\renewcommand{\@makefntext}[1]{%
  \parindent 0pt%
  \noindent\hbox{\@makefnmark}~#1%
}
\makeatother

\title{\LARGE \bf
Learning human-to-robot handovers through 3D scene reconstruction
}

\author{Yuekun Wu, Yik Lung Pang, Andrea Cavallaro, Changjae Oh%
\thanks{This work was supported in part by the CHIST-ERA program through the project CORSMAL, under the UK EPSRC Grant EP/S031715/1 and in part by the Royal Society Research Grant RGS\textbackslash R2\textbackslash242051. 
This research utilised the Sulis Tier 2 HPC platform under the UK EPSRC Grant EP/T022108/1 and the HPC Midlands+ consortium.
}
\thanks{Yuekun Wu, Yik Lung Pang, and Changjae Oh are with Centre for Intelligent Sensing, Queen Mary University of London, UK.
        {\tt\small \{yuekun.wu,y.l.pang,c.oh\}@qmul.ac.uk}. Andrea Cavallaro is with Idiap Research Institute and École Polytechnique Fédérale de Lausanne, Switzerland.
        {\tt\small andrea.cavallaro@epfl.ch}.}%
}

\begin{document}

\maketitle
\thispagestyle{empty}
\pagestyle{empty}

\begin{abstract}
Learning robot manipulation policies from raw, real-world image data requires a large number of robot-action trials in the physical environment. Although training using simulations offers a cost-effective alternative, the visual domain gap between simulation and robot workspace remains a major limitation. Gaussian Splatting visual reconstruction methods have recently provided new directions for robot manipulation by generating realistic environments.
In this paper, we propose the first method for learning supervised-based robot handovers solely from RGB images without the need of real-robot training or real-robot data collection. The proposed policy learner, Human-to-Robot Handover using Sparse-View Gaussian Splatting (H2RH-SGS), leverages sparse-view Gaussian Splatting reconstruction of human-to-robot handover scenes to generate robot demonstrations containing image-action pairs captured with a camera mounted on the robot gripper. As a result, the simulated camera pose changes in the reconstructed scene can be directly translated into gripper pose changes. We train a robot policy on demonstrations collected with 16 household objects and {\em directly} deploy this policy in the real environment. Experiments in both Gaussian Splatting reconstructed scene and real-world human-to-robot handover experiments demonstrate that H2RH-SGS serves as a new and effective representation for the human-to-robot handover task. 

\end{abstract}

\begin{figure*}[t]
  \centering
  \vspace{3 mm}
  \includegraphics[width=1\linewidth]{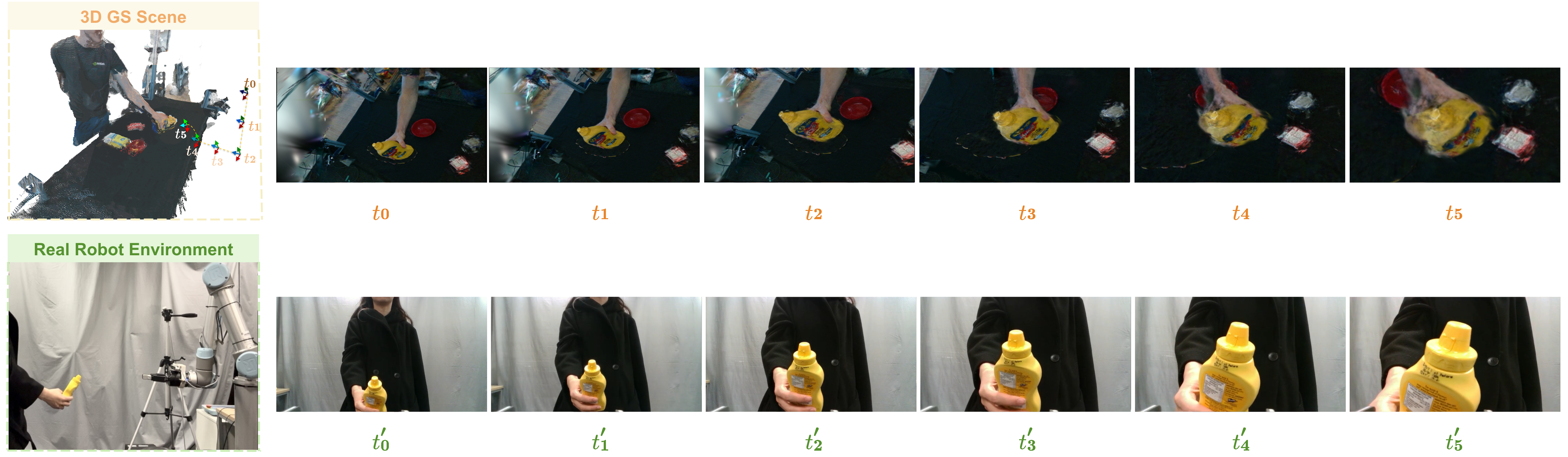}
  \caption{
  Rendered images from a Gaussian Splatting (GS) scene (top) and hand-eye images from a camera in a human-to-robot environment (bottom). In the 3DGS scene, the images are sequentially rendered according to the 6D pose of a virtual hand-eye camera view. In the real robot environment where we perform human-to-robot handovers, our method takes a single image captured by the hand-eye camera on the gripper. 
  Since the movement of the virtual camera in 3DGS scenes can mimic the movement of a hand-eye robot arm in the real robot environment, we train a human-to-robot handover policy from 3DGS scenes and directly deploy this model to the physical human-to-robot environment. $t_0$ ($t'_0$) to $t_5$ ($t'_5$) indicate the sampled time steps when the camera pose is moving from the initial pose to the pre-grasp pose.
  \vspace{-9pt}
  }
  \label{fig0}
\end{figure*}

\section{Introduction}
Training robots to interact with humans in real-world environments presents significant time and safety challenges~\cite{khazatsky2024droid}. While simulations provide a cost-effective alternative to collect training data, they suffer from the domain gap, particularly due to discrepancies in visual appearances and dynamics between simulated and real-world environments~\cite{da2025survey}. To mitigate the domain gap, several approaches focus on improving the rendering of simulated environments to make them more realistic. For example, domain randomization~\cite{domain-random} alters environmental variables, helping the model adapt to diverse real-world conditions. Domain adaptation~\cite{random-adaptation,random-adaptation2} refines the learned policies to better align with real-world data, often using techniques like adversarial learning~\cite{lowd2005adversarial} or feature alignment~\cite{chen2019progressive}. Generative Adversarial Networks are used in domain adaptation to improve the realism of simulated images to facilitate the transfer of skills from simulation to real-world tasks~\cite{GAN,gan2,gan3}. Compared to GAN-based methods, diffusion-based rendering~\cite{diffusion,diff2} generates high-quality images with better generalization, lower data requirements, and more advanced conditioning techniques, such as layout-to-image, text-to-image, and image-to-image generation. However, these methods often require large amounts of data, precise camera calibration, or computationally intensive training~\cite{sim2real}. Additionally, these methods may be restricted to translational motion without rotational adjustments during reaching tasks. In more complex scenarios, these limitations impose additional burdens on learning, which could lead to failure~\cite{RL-GSBridge}. 

Beyond rendering-based solutions, other methods have used images from the real environment to bridge the visual domain gap. Radiance field-based reconstruction methods~\cite{NeRF2Real} capture high-fidelity 3D scene representations, enabling realistic simulation environments for robot training. A sim-to-real method with Gaussian Splatting (GS)~\cite{RL-GSBridge} leverages 3D Gaussian Splatting (3DGS) to construct objects within a simulator, enabling a sim-to-real reinforcement learning framework to address the domain gap. SLAM-based methods~\cite{LearningtoSeebyMoving} construct 3D environment representations through motion and multi-view observations, enhancing spatial understanding and sim-to-real transfer. Additionally, OHPL~\cite{ohpl} learns a policy for robotic reaching movements using a single image captured from the hand-eye camera in the real environment without requiring calibration information or sim-to-real adaptation.
However, these methods are designed for robot manipulation without human interaction.
Hence, they do guarantee human safety during robotic grasping. While several works~\cite{chao2022handoversim,christen2023learning,wang2024genh2r}  proposed simulation environments for human-to-robot handovers using 3D object models and recorded or generated hand and object motion,  simulators do not render realistic images of the environment, thus limiting the input modality of the trained models to point cloud only.

Reconstructing 3D environments for robot training often relies on extensive multi-view datasets, which are costly, time-intensive, and potentially unsafe to acquire. This challenge has spurred interest in efficient learning from sparse camera views, with recent advances in sparse-view GS.  
DNGaussian~\cite{li2024dngaussian}, CoherentGS~\cite{paliwal2025coherentgs}, and FSGS~\cite{zhu2024fsgs} address the challenges of sparse-view reconstruction by leveraging monocular depth estimators to infer depth maps from limited viewpoints. 
Other methods (e.g.~\cite{xu2024mvpgs}) employ multi-view stereo techniques to generate novel-view images as ground truth. These approaches offer a potential solution to the challenge of training robot control models without relying on large-scale multi-camera datasets. However, point clouds derived from RGB images without depth ground-truth often suffer from inaccurate scale estimation.

To address the above-mentioned challenges and facilitate the robot's ability to reach and grasp objects without real training data, we present a novel framework based on hand-eye images changed by a series of discrete robot actions with GS simulations (see Fig.~\ref{fig0}). We hypothesize that, when appropriately formulated, a model trained with GS can effectively learn the necessary skills to perform reaching, grasping and handover tasks within a robot's workspace. To this end, we propose Human-to-Robot Handover using Sparse-View Gaussian Splatting (H2RH-SGS) for human-to-robot handover tasks with reaching and grasping. Unlike existing approaches, H2RH-SGS eliminates the need for real-robot training, human demonstrations, and sim-to-real adaptation. H2RH-SGS derives the robot’s actions directly {\em from a single image} captured by the robot’s hand-eye camera along with the corresponding object and hand masks, enabling handover control through supervised learning. Therefore, the proposed framework effectively bridges the domain gap and addresses the challenges of costly and unsafe data collection with high-fidelity 3DGS for robot action training tasks\footnote{\noindent Code and videos: \url{https://qm-ipalab.github.io/H2RH-SGS/}}. 

\begin{figure*}[t]
  \vspace{2 mm}
  \centering
  \subfigure[Gaussian Splatting reconstruction]{
    \includegraphics[width=0.24\linewidth]{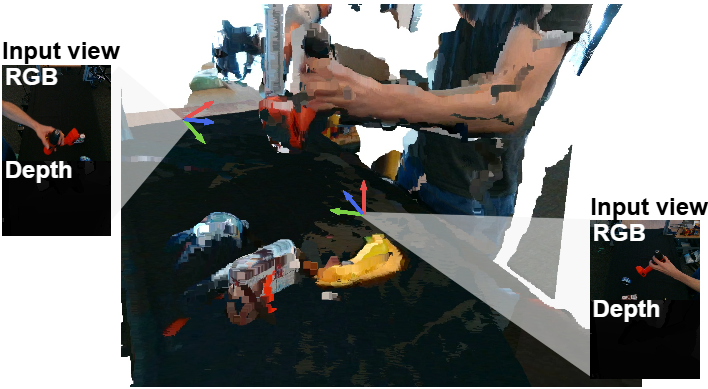}
  }
  \hspace{0.001\linewidth}
  \subfigure[Generating demonstrations]{
    \includegraphics[width=0.24\linewidth]{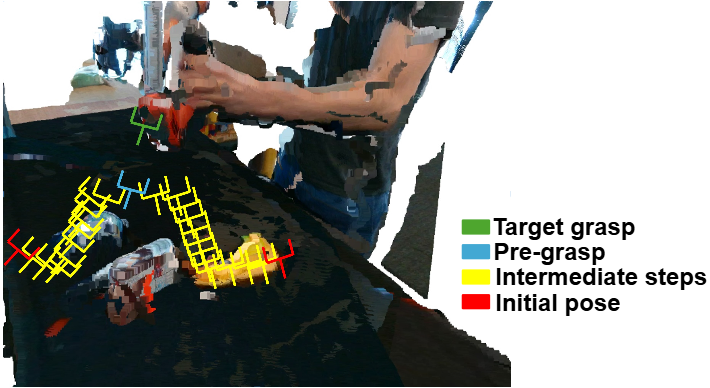}
  }
  \hspace{0.001\linewidth}
  \subfigure[Handover demonstrations]{
    \includegraphics[width=0.18\linewidth]{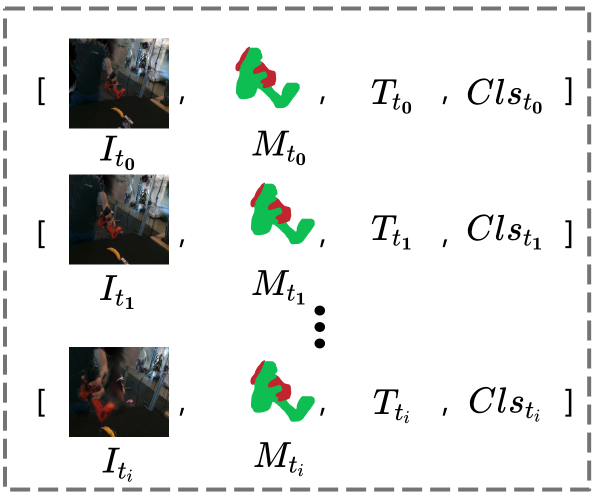}
    }
  \hspace{0.001\linewidth}
  \subfigure[Handover policy training]{
    \includegraphics[width=0.23\linewidth]{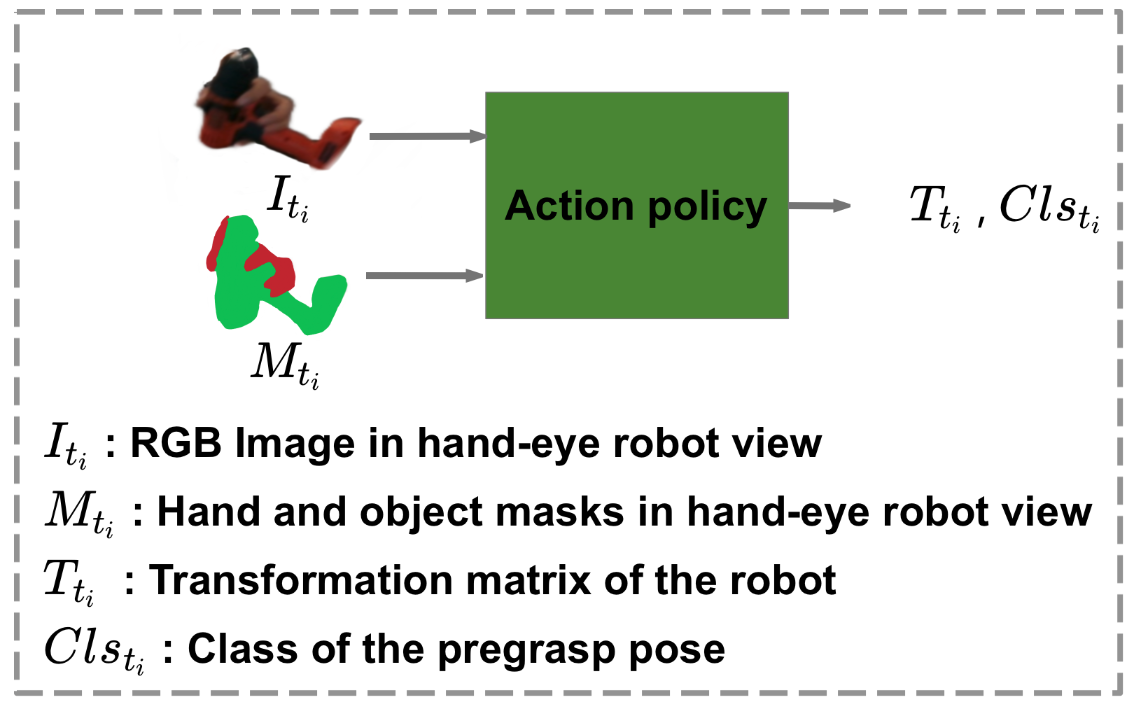}
    }
  \caption{Overview of our method. (a) Given sparse-view RGB-D handover images, we reconstruct a 3D scene using Gaussian Splatting (GS) and then estimate grasp poses from the object and hand point clouds extracted from the GS scene. (b) We then use the GS scene and grasp pose to generate the gripper's trajectory toward the pre-grasp pose and the hand-eye image at each sampled pose. (c) Each trajectory becomes a handover demonstration dataset that includes hand-eye images, object and hand masks, transformations of the gripper pose, and pre-grasp pose labels. (d) The dataset is used to train a handover policy. For inference, only the hand-eye RGB image and masks are required.}
    \label{fig1}
\end{figure*}

\section{Related Work}
This section reviews methods based on GS and learning control policies for robotic manipulation, focusing on reinforcement learning (RL), imitation learning (IL), and hand-eye coordination. We also discuss the differences and contributions of our method.

3DGS~\cite{kerbl20233d} is an explicit radiance field method that directly updates the attributes of each 3D Gaussian component to optimize scene representations. 3DGS uses a splatting technique for rendering and achieves a very high degree of training efficiency. RL-GSBridge~\cite{RL-GSBridge} leverages 3DGS to construct objects within a simulator, enabling a sim-to-real reinforcement learning framework. ManiGaussian~\cite{Manigaussian} employs 3DGS as a dynamic attribute encoder, capturing the spatiotemporal dynamics of the scene to serve as input for reinforcement learning. GraspSplats~\cite{Graspsplats} uses 3DGS to represent and encode grasp-related 3D features, integrating them into an RL framework. Splat-MOVER~\cite{Splat-mover} uses the editability of 3DGS to enable open-vocabulary manipulation tasks. Splat-Nav~\cite{splatnav} proposes a real-time robot navigation pipeline using reconstructed 3D scenes from 3DGS. While these methods primarily use GS to construct scenes that support RL for robot planning and manipulation without human interaction, we leverage GS directly for human-to-robot handover.

Control policies determine the actions of a robot given its current state and environment to achieve specific goals~\cite{andrychowicz2020matters}. 
RL enables robots to acquire control policies through trial-and-error interactions with their environment, optimizing for long-term rewards~\cite{ferede2024end}. RL enables robots to learn end-to-end control through interactions in the real world, but requires extensive training episodes and manual intervention for resetting, thus increasing the cost of data collection~\cite{kalashnikov2018qt, haarnoja2018soft, chebotar2019closing}. To mitigate these challenges, policies are often trained in simulated environments and then transferred to the real world~\cite{tobin2017domain, peng2018sim, random-adaptation2}. However, discrepancies between the dynamics in simulation and the real world, known as the reality gap, remain a significant challenge in RL-based robotic manipulation~\cite{ferede2024end}. IL allows robots to learn from demonstrations provided by human experts or other systems through Behavior Cloning, Generative Adversarial Imitation Learning~\cite{Ho2016GenerativeImitation}, deep imitation learning~\cite{Zhou2020LearningRobots}, 
Inverse Reinforcement Learning~\cite{Ng2000AlgorithmsLearning,Ghosh2019LearningAutonomous}. Expert demonstrations can either directly train the robot or be used to infer reward functions in RL. However, many methods assume demonstrations are expert-optimal~\cite{2}, which is often an overly restrictive condition~\cite{2}: collecting large-scale, high-quality demonstrations is resource-intensive~\cite{115} and often impractical in the real world. Furthermore, human demonstrators are prone to errors, introducing noise into the training data. Finally, most IR methods assume a shared state-action space between the expert and the robot~\cite{106}, restricting their applicability to scenarios where expert demonstrations closely align with the robot's capabilities.

Another related area to our method is hand-eye coordination for robotic policy. A learning-based approach to hand-eye coordination for robotic grasping from monocular images~\cite{hand-eye1} uses 14 robots over a span of two months to collect their dataset and employed visual servoing to generate motion commands. In contrast, our approach does not require real-world robots for data collection, and the network can directly generate the next motion command. OHPL learns a policy for  reach movements using a hand-eye camera without requiring calibration information. However, their method cannot execute grasping tasks, as the robot cannot rotate during its movement. In contrast, our model can perform human-to-robot handover tasks with reaching and grasping in a real-robot environment. HAN~\cite{hand-eyecoordination2} uses a trainable action space capable of approximating human hand-eye coordination behaviors through learning from human-teleoperated demonstrations. However, all tasks are conducted in a simulation environment and the approach has not been implemented on a real robot. Our model, instead, is capable of operating ({\em directly}) on a real robot. ARCap~\cite{chen2024arcap} uses a portable data collection system to gather robot-executable demonstrations and uses augmented reality for visual feedback. Although this approach reduces the collection costs, it still requires people to collect demonstrations.

In summary, while GS methods are commonly employed to reconstruct 3D scenes for RL-based methods, they often suffer from the reality gap and do not focus on human-to-robot handovers. IL methods typically rely on large-scale real-robot expert demonstrations, which are resource-intensive, prone to errors, and costly. Additionally, hand-eye coordination approaches for robotic control often rely on simulated images or necessitate large real-robot datasets, and cannot perform grasping tasks within a real-world robot workspace.
In contrast, our approach directly uses GS for human-to-robot handover, thus bypassing the need for large-scale datasets, real-robot expert demonstrations, calibration information, or sim-to-real adaptation.

\section{Proposed Method}

Fig.~\ref{fig1} shows the main steps of the proposed method. We use sparse-view RGB-D images as input to construct point clouds and process them with GS to reconstruct a 3D scene.
We generate grasp poses by 6-DOF GraspNet~\cite{mousavian20196} with the annotated hand and object mesh.
We then use the reconstructed 3D scene and each grasp pose to generate a camera trajectory towards reaching the pre-grasp pose, which is a pose 30 cm away from the grasp point in the direction of the grasp. Once the robot reaches the pre-grasp pose, the robot enters a closed-loop grasping stage and closes its fingers when the gripper reaches the selected grasp.
Then, we generate the camera trajectories. We sample the initial and intermediate camera poses within the reconstructed 3D scene and render the hand-eye images according to the sampled camera poses. We then use the rendered images and camera poses to construct a \textit{simulated} handover demonstrations dataset, where each data tuple contains a rendered image ($I$), hand and object masks ($M$), a target 6D transformation of the camera ($T\in \mathbb{R}^6$), and a binary label ($Cls\in \{0, 1\}$) indicating if the current state is the pre-grasp pose. Finally, we use this dataset to train a handover policy using a supervised learning approach. During inference, we use a hand-eye RGB image and estimated hand-object masks as input to the trained handover policy to predict the transformation of the hand-eye robot and classify if the robot has reached the pre-grasp pose.

\subsection{Grasp pose estimation in 3D scenes}\label{sec_grasp_pose}
We reconstruct a 3D scene using GS and then create grasp poses for the object within this scene. Given sparse-view RGBD images of a human holding an object, we aim to reconstruct the 3D scene in metric scale as a simulation environment for robot learning with real point clouds distance units and realistic visuals. 
Since GraspNet uses a metric scale for grasp pose prediction, we need the point clouds to be in metric scale for proper alignment with the grasp pose. 
By reconstructing the scene in metric scale, the transformations of the camera in the reconstructed scene can be directly translated to the robot gripper transformations in the real environment.
We use FSGS to reconstruct the 3D scene from sparse-view RGBD images. Since FSGS is initialized from COLMAP, the resulting point clouds are not in metric scale. As a result, the reconstructed point cloud cannot be directly aligned with the grasp poses from GraspNet. Furthermore, the camera transformations in the reconstructed scene cannot be directly translated to the robot gripper’s movements in the real environment.
Hence, we use depth maps to initialize the point clouds and obtain metric scale and high-quality 3D reconstruction results.

Next, we generate the target grasp pose by selecting a diverse set of grasp candidates on the target object. To generate grasp poses, the annotated point clouds of the object, \( P_{\text{o}} \), and the hand, \( P_{\text{h}} \), from the dataset are extracted and processed by GraspNet. We select the grasp with the highest score as the grasp pose,  \( G_{\text{o}} \in SE(3) \).

To ensure safety, we check whether a grasp pose collides with the human hand by transforming the hand point cloud into the grasp coordinate frame and computing the Euclidean distance:
\begin{equation}
   {p}_{\text{h}}' = {G_{\text{o}}}^{-1} {p}_{\text{h}},
\end{equation}
where \( {G}_{\text{o}}\) is the grasp pose, \(p_{\text{h}} \in {R}^3\) is a point from the hand point cloud. A grasp pose is considered unsafe if a point from the hand is within the safety threshold \(d_{\text{s}}\). In such a case the grasp is removed to prevent harm to the human.

For the coordinate system of the estimated grasps, the origin is located at the center of the object, and the axes are aligned with the camera. Therefore, we need to align the estimated grasp with the reconstructed 3D scene as follows:
\begin{equation}
G_\text{GS} \ = G_{\text{o}} - \text{Mean}_{\text{o}},
\end{equation}
where \(G_{\text{GS}} \in SE(3)\) is the grasp pose in the reconstructed 3D scene, \( G_{\text{o}} \) is the grasp pose in the object coordinate system, and \( \text{Mean}_{\text{o}} \) is the mean position of the object in the reconstructed 3D scene.

\subsection{Handover demonstrations dataset}\label{sec_handover_dataset}
Next, we construct a dataset of simulated robot demonstrations for handovers, which consists of a tuple $(I, M, T, Cls)$ for each time step. 
First, we sample \( k\) initial poses for each target grasp $G_\text{GS}$. For initial pose sampling, candidate poses \( q_i \in SE(3)\) are generated using spherical sampling around the grasp pose \( G_{\text{GS}} \) with radius \( r \):
\begin{equation}
\|G_\text{GS} - q_\text{i}\| = r,
\end{equation}

Then, we establish a series of filtering mechanisms for selecting initial points. To ensure the plausibility of the trajectory, the initial camera position $T_i\in \mathbb{R}^3$ and hand position $T_\text{h}\in \mathbb{R}^3$ should be on different sides of the object centered at \(T_{\text{o}}\in \mathbb{R}^3\):
\begin{equation}
\arccos \left( \frac{(T_{\text{h}} - T_{\text{o}}) \cdot (T_i - T_{\text{o}})}{\|T_{\text{h}} - T_{\text{o}}\| \cdot \|T_i - T_{\text{o}}\|} \right) > \alpha_{\text{min}},
\end{equation}
where \(\alpha_{\text{min}}\) is the minimum angle to ensure that the hand and initial camera are on opposite sides of the object. 

To ensure diversity and generalization in sampling, we sample the initial poses with a random angular offset, while setting the sampled poses to generally face the object:
\begin{equation}
z_{\text{i}} = R_x(\theta_x) \cdot R_y(\theta_y) \cdot\frac{T_{\text{o}} - T_{\text{i}}}{\|T_{\text{o}} - T_{\text{i}}\|},
\end{equation}
where \(z_{\text{i}}\) is the $z$-axis of the initial sampling point, \(R_x(\theta_x) \in SO(3) \) and \(R_y(\theta_y) \in SO(3)\) are the rotation matrix around the $x$-axis and $y$-axis, respectively. 

To avoid generating unreasonable trajectories, we formulate the constraint that the angle between the initial sampling point and the final grasp point does not exceed a given  angle \(\theta_{\text{max}}\):
\begin{equation}
\cos \theta = z_{\text{i}} \cdot z_{\text{f}} \geq \cos(\theta_{\text{max}}),
\end{equation}
where \(z_{\text{f}}\) is the $z$-axis direction of the final grasp point and \(\theta_{\text{max}}\) is the maximum angle. 
\begin{figure}[t]
  \centering
  \subfigure{
    \includegraphics[width=0.33\linewidth]{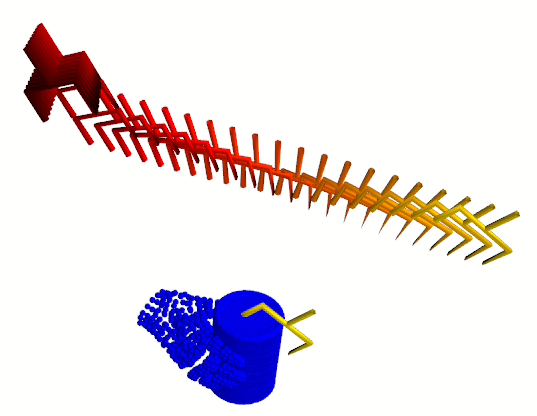}
  }
  \hspace{0.1\linewidth}
  \subfigure{
    \includegraphics[width=0.3\linewidth]{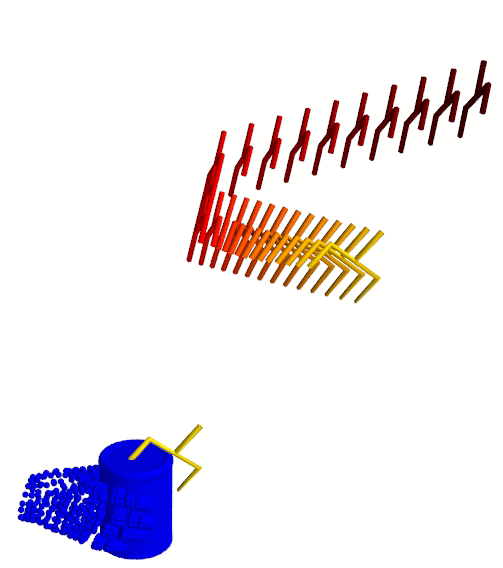}
  }
  \caption{Examples of sampling trajectories from the initial pose to the pre-grasp pose. The gripper's trajectories are color-coded from dark red to yellow to represent the temporal evolution of actions. The blue point clouds represent the hand and object. The yellow gripper on the point cloud indicates the grasp pose.}
  \label{grasp-poses}
\end{figure}

Next, we generate trajectories from the initial pose to the pre-grasp pose. Fig.~\ref{grasp-poses} shows examples of sampling trajectories from the initial pose to the pre-grasp pose. We generate one trajectory for each initial pose, and hence we have \( k\) trajectories for each target grasp $G_\text{GS}$.
For each target grasp $G_\text{GS}$, we compute a pre-grasp pose \(T_\text{pre}\in \mathbb{R}^3\) which is positioned \(s\) meters away from the grasp pose along the grasp pose $z$-axis vector \( \hat{n} \):
\begin{equation}
T_\text{pre} =  T_{\text{GS}}  - s \cdot \hat{n},
\end{equation}
where \(T_{\text{GS}}\in \mathbb{R}^3\) is the grasp position, and \(\hat{n}\in \mathbb{R}^3\) is the unit vector along grasp pose's $z$-axis.

The transition from the initial sampled point to pre-grasp pose involves three phases: In the first phase, the robot will move to a position where the gripper's reaching direction is aligned with the object, ensuring that the object appears at the center of the sampled image. We denote the time at which the first phase ends as \(t_1\).

During the second phase, only the position is adjusted until the distance to the pre-grasp pose is less than a threshold \(d\). The position transition is performed with linear interpolation:
\begin{equation}
\|T - T_\text{pre}\| < d,
\label{eq:distance_threshold}
\end{equation}
\begin{equation}
T(\tau) = (t_2 - \tau) \cdot T_{1} + \tau \cdot T_{2}, \quad \tau \in [t_1, t_2].
\end{equation}
We denote the last time of the second stage, at which Eq.~\eqref{eq:distance_threshold} reaches the threshold, as \(t_2\). \(T(\tau)\in \mathbb{R}^3\) is the position at time \(\tau\). \(T_{1}\in \mathbb{R}^3\) is the position at the last moment of phase 1. \(T_{2}\in \mathbb{R}^3\) is the position at the last moment of phase 2. 

In the third phase, both position and rotation are refined until the pre-grasp pose is reached. The movement is performed using a combination of linear interpolation for position and Spherical Linear Interpolation (SLERP) for rotation:
   \begin{equation}
   T(\tau) = (t_{pre} - \tau) \cdot T_\text{2} + \tau \cdot T_\text{pre}, \quad \tau \in [t_2, t_{pre}],
   \end{equation}
   \begin{equation}
    R(\tau) = \text{SLERP}(R_{\text{i}}, R_{\text{pre}}, \tau), \quad \tau \in [t_2, t_{pre}],
   \end{equation}
   where \(R(\tau)\in SO(3)\) is the rotation at time \(\tau\),  \(R_\text{i}\in SO(3)\) is the rotation at the initial pose, and  \(R_\text{pre}\in SO(3)\) is the rotation at the pre-grasp pose.

We set the gripper's reaching direction to be aligned with the object by formulating the robot facing the object as:
    \begin{equation}
    \theta > \varphi \implies R_{\text{update}},
    \end{equation}
    \begin{equation}
    R(\tau) = \text{SLERP}(R_{\text{update}}, R_{\text{pre}}, \tau),
    \end{equation}
    where \(\theta\) is the angle between the robot's current orientation and the direction to the object, \(\varphi\) is the threshold for angle change, and \(R_{\text{updated}}\in SO(3)\) is the updated rotation matrix that aligns the robot to face the object, and \(R_{\text{pre}} \in SO(3)\) is the pre grasp pose's rotation matrix.

To ensure safety, the poses must maintain sufficient distance from the hand and the object point cloud:
\begin{equation}
\min_{p \in P_{\text{o}} \cup P_{\text{h}}} \|T_i - p\| > d_{\text{min}},
\end{equation}
where \( P_{\text{o}} \cup P_{\text{h}} \) represents the point cloud of the hand and the object, and \( d_{\text{min}} \) is the minimum allowable distance between gripper and the hand and object.

Finally, at each timestep we render the hand-eye camera view given the 6D pose of the hand-eye camera, \( T \in SE(3)\), and the reconstructed scene. The 3D Gaussians are projected into the 2D image space, and we use fast $\alpha$-blending~\cite{kerbl20233d} to render the hand-eye image efficiently. At the same time, we use \( T \)\ to project the hand and object point clouds to 2D, obtaining the hand and object masks.

\subsection{Policy learning for human-to-robot handover}\label{sec_policy}
We use the simulated handover dataset to train a policy network for robot control. We design a policy network that predicts the robot's translational and rotational movements while determining whether a grasp action should be executed at a given moment. The policy network uses an RGB hand-eye robot image $I$ without a background and a hand-object mask $M$ as input, and outputs the transformed pose $T'\in \mathbb{R}^6$  and grasping-decision $C\in [0, 1]$. The mathematical formulation is as follows:
\begin{equation}
T',C= Policy(I,M), \\
\end{equation}
where $I$ represent hand-eye robot image without background, $M$ represents the mask of the hand and object, $T'$ is the predicted transformation pose, and $C$ is a continuous value representing the predicted classification indicating whether the current pose is a pre-grasp pose.

The output vector $T'$ consists of $(T'_\text{rot}, T'_\text{trans})$, where $T'_\text{rot}$ represents the 3D rotation in Euler angles, and $T'_\text{trans}$ represents the translation. Next, we convert \( T' \) into the \( T \in SE(3)\), which serves as the transformed pose output of the policy.
 The binary classification output $Cls$ is defined as:
\begin{equation}
    Cls =
    \begin{cases}
        1, & \text{if } C C \geq \tau_c, \\
        0, & \text{otherwise}.
    \end{cases}
\end{equation}
When \( C\) exceeds a pre-defined threshold $\tau_c$, we consider that the robot has reached a pre-grasp pose (\( Cls = 1 \)). 
Otherwise, the robot continues to reach the object until one of the following conditions is met: either the number of time steps reaches 30 in the simulation, or the reaching time in the real robot workspace reaches 30 seconds.

The total loss function $L$ consists of the translation loss \( L_{\text{T}} \), the rotation loss \( L_{\text{R}} \), and the classification loss \( L_{\text{C}} \), each of which is optimized using the Mean Squared Error (MSE) loss as:
\begin{equation}
L = \lambda_{\text{T}} L_{\text{T}} + \lambda_{\text{R}} L_{\text{R}} +  L_{\text{C}},
\end{equation}
where the terms \( \lambda_{\text{T}} \) and \( \lambda_{\text{R}} \) are hyperparameters for weighting \( L_{\text{T}} \) and \( L_{\text{R}} \), respectively.

\section{Experiments}
We validate our method through both simulations in a reconstructed 3D scene and real-robot experiments. In the 3D scene, we simulate the robot reaching the pre-grasp pose using an action policy that predicts the transformation matrix based on an RGB image and object mask, and we analyze both quantitative and visual results. For real-robot validation, we conduct handover experiments, evaluating the success rate, safety, and time taken to reach the pre-grasp pose. We compare our method to an image-based visual servoing algorithm.

\subsection{Setup}
\begin{figure}[t]
  \vspace{2 mm}
  \centering  \includegraphics[width=0.97\linewidth]{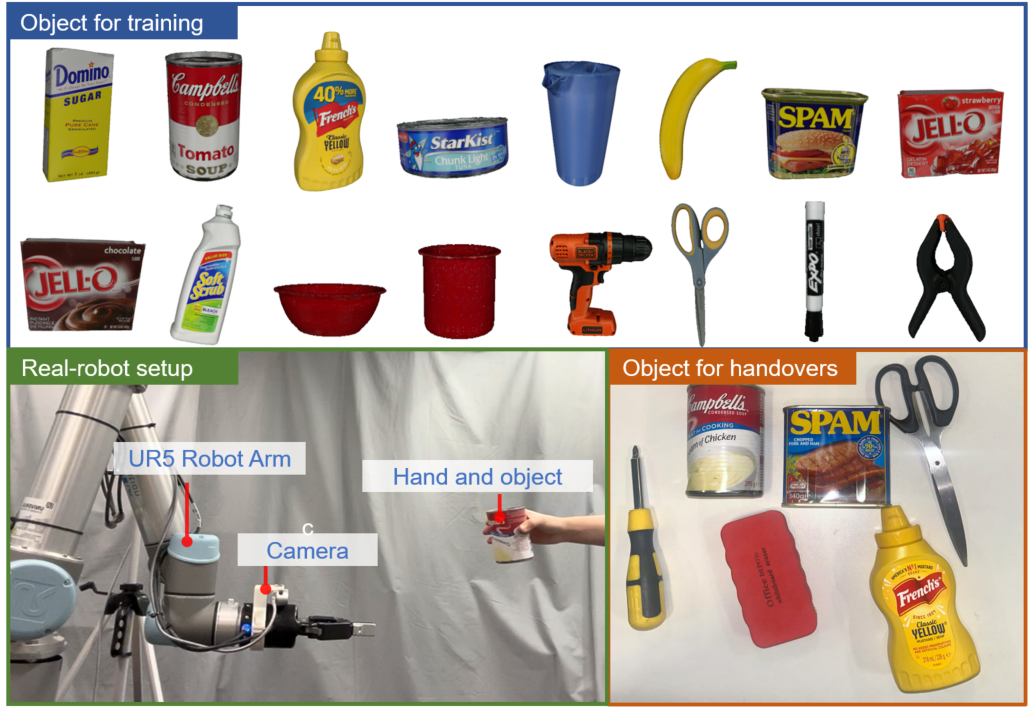}
  \caption{(Top) 16 household objects for policy training. (Bottom-left) Real-robot experiment Setup: UR5 robotic arm with Robotiq 2F-85 2-finger gripper and hand-eye Intel Realsense D435i camera, and a participant holding a object. (Bottom-right) 6 test objects (4 seen/2 unseen during training) in real-robot handover experiments.}  
  \label{SETUP}
\end{figure}
\subsubsection{Parameter settings} 
In grasp pose estimation (Sec.~\ref{sec_grasp_pose}), we set the safety threshold for the grasp pose to \(d_{\text{s}} = 0.1 \) meters from the human. 
For constructing the handover demonstrations dataset (Sec.~\ref{sec_handover_dataset}), we set \( k =15\) for each target grasp $G_\text{GS}$, which is consistent with data collection in real robot environments, and $r = 0.7$, which aligns with the typical initial distance for the UR5 robotic arm. 
To generate reasonable handover trajectories, we set \(\alpha_{\text{min}} = 60 ^\circ\), random values within \([-20^\circ, 20^\circ]\) for $\theta_x$ and $\theta_y$, and \(\theta_{\text{max}} =100 ^\circ\).
In trajectory generation, we set \(s=0.3\), \( d = 0.5 \), \(\varphi = 0\), and \( d_{\text{min}} = 0.1 \) to be compatible with our real-robot environment. 
Finally, in policy learning (Sec.~\ref{sec_policy}), we set $\tau_c = 0.7$ in simulation, $\tau_c = 0.6$ in the real-robot environment, and $\lambda_{\text{T}}=100, \lambda_{\text{R}}=100$ for loss functions.

\subsubsection{Robot reaching simulation}
We validate the action policy in the reconstructed 3D scenes by guiding the simulated hand-eye robot to reach the pre-grasp pose without harming the human hand.
Given an RGB image and mask containing only the hand and object without the background, the action policy predicts the next transformation matrix \( T\), and \( Cls\). Using GS, we can render the next hand-eye view image at \( T \) and this process is repeated until the action policy outputs a pre-grasp pose classification \( Cls =1\), or the number of time steps reaches 30 in the reconstructed 3D scene, or the reaching time in the real robot workspace reaches 30 seconds.
Experiments are conducted on all 11 objects, with each object having 10 starting positions.

\subsubsection{Real-robot handover}
As shown in Fig.~\ref{SETUP}, the setup includes a 6-DoF Universal Robots UR5 robotic arm, equipped with a Robotiq 2F-85 2-finger gripper for grasping the object, and an Intel RealSense D435i camera fixed on the robot arm.
The protocol configurations are executed by 4 participants, with each participant standing stationary throughout all experiments, and holding the object stationary. For each object, experiments were conducted at five different positions: middle, left, right, top, and bottom. The participants were instructed to grasp the lower half of the object. We used 6 objects with diverse shapes and textures, some of which were seen during training while others were entirely unseen. The testing environment was distinct from the training environment.
To generate hand and object masks, we first estimate the bounding boxes of both the hand and the (grasped) object with a hand-object detector~\cite{yik32}. 
Within the object bounding box, we segment the object via Fast Segment Anything~\cite{yik33}. The hand region is cropped and processed by FrankMocap~\cite{yik34} to extract a hand mask from the predicted silhouette.
For real-robot validation, we define successful grasping based on three criteria: 
\begin{itemize}[left=0em]
  \item \textit{Grasp success rate}: The percentage of attempts where the robot successfully grasps and secures the target object.
  \item \textit{Safety rate}: The proportion of safe grasps that do not harm the human hand, within successful grasps.
  \item \textit{Elapsed time}: The duration measured from the robot's initial position to reaching the pre-grasp pose, quantifying the efficiency of the grasping process.
\end{itemize}
As we are the first to propose a method that learns human-to-robot handover solely from RGB images without real-robot training or real-robot data, there is no direct baseline for comparison. Since our task is similar to image-based visual servoing (IBVS), we implemented an algorithm for IBVS, which utilizes object masks to guide the robot's handover task (IBVS-MASK). IBVS-MASK estimates the mask centroid and adjusts the robot position to align the mask centroid with the image center. Once the mask's area exceeds a predefined threshold relative to the total image area, the robot is considered to have reached the pre-grasp pose and the grasping action is executed.

\subsection{Analysis}
\subsubsection{Simulation results in the reconstructed 3D scenes}

\begin{table}[t]
    \vspace{3 mm}
    \centering
    \caption{Performance of reaching to pre-grasp attempts across multiple trials per object.}
    \label{gs-table}
    \renewcommand{\arraystretch}{1.0} 
    \setlength{\tabcolsep}{1.2pt}    
    \begin{tabular}{l l ccc} 
        \hline
        \textbf{ } & 
        \textbf{Object} & 
        \textbf{Mean Dis $\pm$ Std} & 
        \textbf{Safe Rate} & 
        \textbf{Center Rate} \\
        \hline
        \multirow{9}{*}{\rotatebox[origin=c]{0}{\textbf{Seen}}}
        & Mustard Bottle & 0.37 $\pm$ 0.10 & 10/10 & 10/10 \\
        & Tomato Soup & 0.44 $\pm$ 0.01 & 10/10 & 10/10 \\
        & Potted Meat & 0.36 $\pm$ 0.05 & 10/10 & 10/10 \\
        & Blue Kettle & 0.44 $\pm$ 0.04 & 10/10 & 10/10 \\
        & Gelatin Box & 0.44 $\pm$ 0.04 & 10/10 & 10/10 \\
        & Red Bowl & 0.52 $\pm$ 0.04 & 10/10 & 10/10 \\
        & Scissors & 0.25 $\pm$ 0.03 & 10/10 & 10/10 \\
        & Banana & 0.35 $\pm$ 0.01 & 9/10 & 10/10 \\
        & Power Drill & 0.38 $\pm$ 0.01 & 10/10 & 10/10 \\
        \hline
         \multirow{2}{*}{\rotatebox{0}{\textbf{Unseen}}}
        & Master Chef & 0.46 $\pm$ 0.04 & 8/10 & 8/10 \\
        & Cracker Box & 0.37 $\pm$ 0.19 & 8/10 & 8/10 \\
        \hline
    \end{tabular}
\end{table}
\begin{figure}[t]
  \centering
  \subfigure{%
    \includegraphics[width=0.98\linewidth]{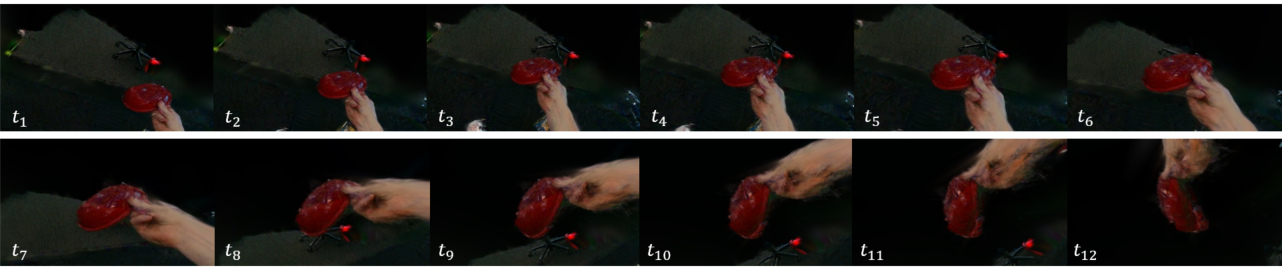}%
  }%
  
  \subfigure{%
    \includegraphics[width=0.99\linewidth]{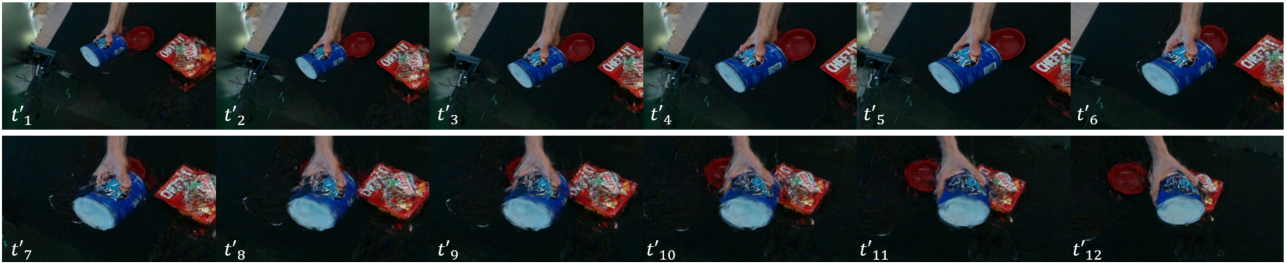}%
  }%
  \caption{Trajectory visualization in the 3D scene, showing paths from initial pose to pre-grasp pose. (First two rows) Trajectory  for an object that was seen during training. (Last two rows) Trajectory for a novel object, which was not seen in training.}
  \label{render-vis}
  \vspace{-9pt}
\end{figure}
\begin{table*}[t]
    \vspace{3 mm}
    \centering
    \caption{Comparison between the image-based visual servoing method (IBVS-MASK) and ours for human-to-robot handovers in a real robot workspace. Metrics include success rate ($\uparrow$), safety rate ($\uparrow$), and time taken from initial pose to pre-grasp pose ($\downarrow$).}
    \label{real}
    \vspace{-5pt}
    \renewcommand{\arraystretch}{1.0} 
    \setlength{\tabcolsep}{1.2pt}       
    \begin{tabular}{l ccc ccc ccc ccc ccc ccc}
        \hline
        \multirow{2}{*}{\textbf{Model}} 
        & \multicolumn{3}{c}{\textbf{Mustard Bottle}}  
        & \multicolumn{3}{c}{\textbf{Potted Meat Can}} 
        & \multicolumn{3}{c}{\textbf{Scissors}} 
        & \multicolumn{3}{c}{\textbf{Tomato Soup Can}} 
        & \multicolumn{3}{c}{\textbf{Blackboard Eraser}} 
        & \multicolumn{3}{c}{\textbf{Screwdrivers}} \\ 
        \cline{2-19}
        & Success & Safe & Time (s) 
        & Success & Safe & Time (s) 
        & Success & Safe & Time (s) 
        & Success & Safe & Time (s) 
        & Success & Safe & Time (s) 
        & Success & Safe & Time (s) \\ 
        \hline
        IBVS-MASK & 0.33 & 0.60 & 18.12 & 0.35 & \textbf{1.00}&	21.16 &0.05 &	\textbf{1.00} &	23.00 & 0.40&	0.87&	14.86 & 0.20 &	0.33 &	\textbf{7.80} & 0.05 &	\textbf{1.00} &	 \textbf{7.30}
        \\
        Ours& \textbf{0.93} & \textbf{1.00} & \textbf{10.45} & \textbf{0.85} &  \textbf{1.00}& \textbf{12.75} & \textbf{0.35} & \textbf{1.00} & \textbf{9.65} & \textbf{0.55} & \textbf{1.00} & \textbf{11.11} & \textbf{0.65} & \textbf{0.84} & 9.29 & \textbf{0.70} & \textbf{1.00} & 10.13 \\
        \hline
    \end{tabular}
\end{table*}
Table~\ref{gs-table} presents the  results of the simulated robot reaching the pre-grasp point in the 3D scene. For validation in the 3D scene, a successful reach to the pre-grasp point is defined as follows: the robot maintains a distance from the object between 0.35 and 0.45 meters. The robot's gripper faces the object, meaning that the object is positioned at the center of the hand-eye image;  Additionally, the robot ensures a safe interaction by moving in a straight line to safely grasp the object without harming the human.
The table presents three metrics: the distance between the simulated robot and the object at the pre-grasp pose, the object center rate in the hand-eye camera view, and the safety rate of the grasp attempt.
The results indicate that our model’s predicted pre-grasp poses consistently maintain a distance of approximately 0.4 meters from the object, aligning well with the distance of the designed trajectory. Furthermore, the pre-grasp poses are generally oriented towards the center of the object, allowing the robot to successfully grasp it by moving directly along a straight path after reaching the pre-grasp pose. And most of the learned pre-grasp poses avoid touching areas where hand are, ensuring handover safety.

Fig.~\ref{render-vis} shows visual results of the simulated robot approaching the object in the 3D scene. For both seen and unseen objects, the model consistently reaches the pre-grasp pose. The simulated robot always first moves to face the object, then approaches it, and finally rotates into an optimal pre-grasp pose for grasping. These results show the model's capability to achieve the pre-grasp pose in the 3D scene, consistent with our intended trajectory design.
\begin{figure}[t]
  \centering
  \includegraphics[width=0.98\linewidth]{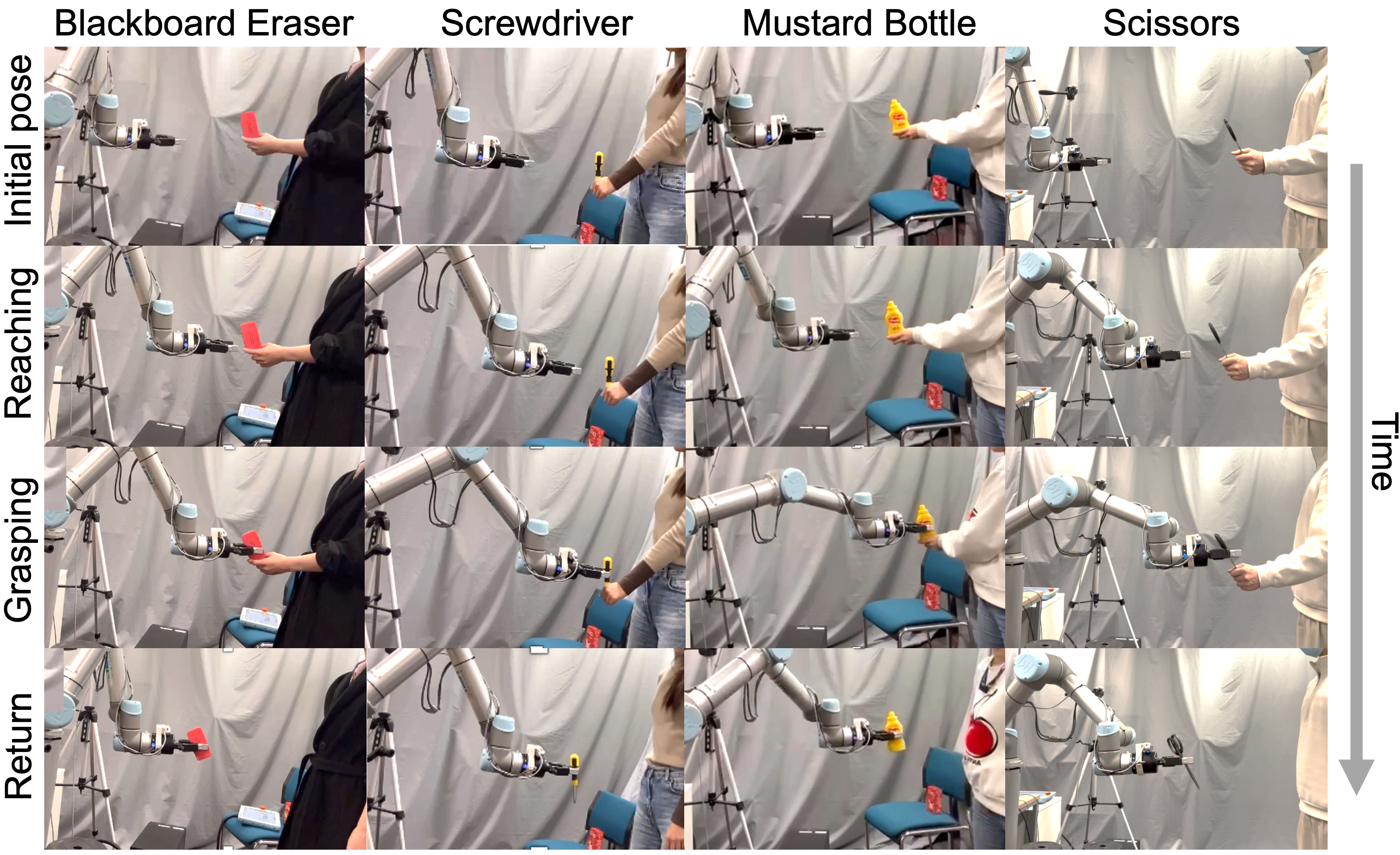}
  \caption{Examples of successful human-to-robot handovers of household objects using H2RH-SGS with RGB image and hand-object mask.}
  \label{fig5}
  \vspace{-9pt}
  \vspace{-9pt}
\end{figure}
\subsubsection{Real-robot validation}
In Table~\ref{real}, we compare our model's handover performance with IBVS-MASK in real environments. In Fig.~\ref{fig5}, we showed that our method can perform human-to-robot handovers on general household objects. As observed from the results, our model exhibits robust and effective handover performance for both seen and unseen objects. Compared to IBVS-MASK, our approach achieves higher grasping success rates, higher safety rate, and reduces execution time. IBVS-MASK struggles with thin objects such as scissors, blackboard erasers, and screwdrivers, failing to accurately identify object centers and reach the predetermined mask-to-image ratio threshold. This limitation impairs IBVS-MASK's ability to approach and grasp these object types effectively. In contrast, our method overcomes these challenges, exhibiting robust performance across a diverse range of objects, including thin items, showing broader applicability. Furthermore, IBVS-MASK's reliance on object mask centroids for robot control often results in oscillatory movements during the reaching phase, leading to prolonged grasping times. Our method eliminates this problem, enabling smooth and efficient object approaches. These results show that our proposed method can perform handover tasks safely, robustly, and efficiently, with smooth execution. 

\section{Conclusion}
We proposed the first method for learning supervised-based robot handovers solely from RGB images without the need of real-robot training or collecting real-robot data. The key idea is the generation of robot demonstrations in a reconstructed 3D scene to simulate hand-eye image view changes through a series of discrete robot actions. Next, we use the demonstration data to train a policy model for human-to-robot handover. We demonstrate the feasibility of our method in both reconstructed 3D scenes and a real-world environment. Future work includes incorporating hand-related features~\cite{liang2024vton} to better guide rotational movements of the robot and leveraging world foundation models~\cite{agarwal2025cosmos} to generate more plausible and realistic demonstration data.

\addtolength{\textheight}{-3.5cm}

{
\bibliographystyle{IEEEtran} 
\bibliography{main}
}

\end{document}